\documentclass[11pt]{article}

\usepackage{acl2014}
\usepackage{times}
\usepackage{url}
\usepackage{latexsym}
\usepackage{etoolbox}

\usepackage{graphicx}
\usepackage{url}
\usepackage{amsmath}
\usepackage{amssymb}
\usepackage{color}
\usepackage{caption}
\usepackage{multirow}
\usepackage{comment}
\usepackage[utf8]{inputenc}

\newcommand{\bmx}[0]{\begin{bmatrix}}
\newcommand{\emx}[0]{\end{bmatrix}}
\newcommand{\qt}[1]{\left<#1\right>}

\newcommand{\vect}[1]{\mathbf{#1}}

\newcommand{\matr}[1]{\mathbf{#1}}

\newcommand{\vh}[0]{\vect{h}}

\newcommand{\vx}[0]{\vect{x}}
\newcommand{\vw}[0]{\vect{w}}

\newcommand{\vr}[0]{\vect{r}}

\newcommand{\vz}[0]{\vect{z}}

\newcommand{\mW}[0]{\matr{W}}
\newcommand{\mG}[0]{\matr{G}}

\newcommand{\mU}[0]{\matr{U}}

\newcommand{\RR}[0]{\mathbb{R}}

\graphicspath{ {./figures/} }

\title{On the Properties of Neural Machine Translation: Encoder--Decoder Approaches}

\author{
    Kyunghyun Cho~~~~~~~~~Bart van Merri\"enboer\\
    Universit\'e de Montr\'eal \\
  \And
  Dzmitry Bahdanau\thanks{\hspace{2mm}Research done while visiting Universit\'e de
    Montr\'eal}\\
    Jacobs University, Germany \\
  \AND
    Yoshua Bengio \\
    Universit\'e de Montr\'eal, CIFAR Senior Fellow
}

\date{}

\begin{document}
\maketitle

\begin{abstract}
    Neural machine translation is a relatively new approach to statistical
    machine translation based purely on neural networks. The neural machine
    translation models often consist of an encoder and a decoder. The encoder
    extracts a fixed-length representation from a variable-length input
    sentence, and the decoder generates a correct translation from this
    representation. In this paper, we focus on analyzing the properties of the
    neural machine translation using two models; RNN Encoder--Decoder and a
    newly proposed gated recursive convolutional neural network. We show that
    the neural machine translation performs relatively well on short sentences
    without unknown words, but its performance degrades rapidly as the length of
    the sentence and the number of unknown words increase. Furthermore, we find
    that the proposed gated recursive convolutional network learns a grammatical
    structure of a sentence automatically.
\end{abstract}

\section{Introduction}

A new approach for statistical machine translation based purely on neural
networks has recently been proposed~\cite{Kalchbrenner2012,Sutskever2014}. This
new approach, which we refer to as {\it neural machine translation}, is inspired
by the recent trend of deep representational learning. All the neural network
models used in \cite{Kalchbrenner2012,Sutskever2014,Cho2014} consist of an
encoder and a decoder. The encoder extracts a fixed-length vector representation
from a variable-length input sentence, and from this representation the decoder
generates a correct, variable-length target translation.

The emergence of the neural machine translation is highly significant, both
practically and theoretically. Neural machine translation models require only a
fraction of the memory needed by traditional statistical machine translation
(SMT) models. The models we trained for this paper require only 500MB of memory
in total.  This stands in stark contrast with existing SMT systems, which often
require tens of gigabytes of memory. This makes the neural machine translation
appealing in practice. Furthermore, unlike conventional translation systems,
each and every component of the neural translation model is trained jointly to
maximize the translation performance.

As this approach is relatively new, there has not been much work on analyzing
the properties and behavior of these models. For instance: What are the
properties of sentences on which this approach performs better? How does the
choice of source/target vocabulary affect the performance? In which cases does
the neural machine translation fail?

It is crucial to understand the properties and behavior of this new neural
machine translation approach in order to determine future research directions.
Also, understanding the weaknesses and strengths of neural machine translation
might lead to better ways of integrating SMT and neural machine  translation
systems.

In this paper, we analyze two neural machine translation models. One of them is
the RNN Encoder--Decoder that was proposed recently in \cite{Cho2014}. The other
model replaces the encoder in the RNN Encoder--Decoder model with a novel neural
network, which we call a {\it gated recursive convolutional neural network}
(grConv). We evaluate these two models on the task of translation from French to
English.

Our analysis shows that the performance of the neural machine translation model
degrades quickly as the length of a source sentence increases. Furthermore, we
find that the vocabulary size has a high impact on the translation performance.
Nonetheless, qualitatively we find that the both models are able to generate
correct translations most of the time. Furthermore, the newly proposed grConv
model is able to learn, without supervision, a kind of syntactic structure over
the source language.

\section{Neural Networks for Variable-Length Sequences}

In this section, we describe two types of neural networks that are able to
process variable-length sequences. These are the recurrent neural network
and the proposed gated recursive convolutional neural network.

\subsection{Recurrent Neural Network with Gated Hidden Neurons}
\label{sec:rnn_gated}

\begin{figure}[ht]
    \centering
    \begin{minipage}{0.44\columnwidth}
        \centering
        \includegraphics[width=0.8\columnwidth]{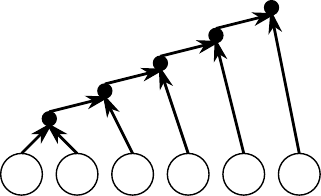}
    \end{minipage}
    \hfill
    \begin{minipage}{0.44\columnwidth}
        \centering
        \includegraphics[width=0.8\columnwidth]{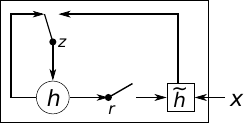}
    \end{minipage}
    \medskip
    \begin{minipage}{0.44\columnwidth}
        \centering
        (a)
    \end{minipage}
    \hfill
    \begin{minipage}{0.44\columnwidth}
        \centering
        (b)
    \end{minipage}
    \caption{The graphical illustration of (a) the recurrent
        neural network and (b) the hidden unit that adaptively forgets and remembers.}
    \label{fig:rnn_unit}
\end{figure}

A recurrent neural network (RNN, Fig.~\ref{fig:rnn_unit}~(a)) works on a variable-length sequence $x=(\vx_1,
\vx_2, \cdots, \vx_T)$ by maintaining a hidden state $\vh$ over time. At each
timestep $t$, the hidden state $\vh^{(t)}$ is updated by
\begin{align*}
    \vh^{(t)} = f\left( \vh^{(t-1)}, \vx_t \right),
\end{align*}
where $f$ is an activation function. Often $f$ is as simple as performing a
linear transformation on the input vectors, summing them, and applying an
element-wise logistic sigmoid function.

An RNN can be used effectively to learn a distribution over a variable-length
sequence by learning the distribution over the next input $p(\vx_{t+1} \mid
\vx_{t}, \cdots, \vx_{1})$. For instance, in the case of a sequence of
$1$-of-$K$ vectors, the distribution can be learned by an RNN which has as an
output
\begin{align*}
    p(x_{t,j} = 1 \mid \vx_{t-1}, \dots, \vx_1) = \frac{\exp \left(
        \vw_j \vh_{\qt{t}}\right) } {\sum_{j'=1}^{K} \exp \left( \vw_{j'}
        \vh_{\qt{t}}\right) },
\end{align*}
for all possible symbols $j=1,\dots,K$, where $\vw_j$ are the rows of a
weight matrix $\mW$. This results in the joint distribution
\begin{align*}
    p(x) = \prod_{t=1}^T p(x_t \mid x_{t-1}, \dots, x_1).
\end{align*}

Recently, in \cite{Cho2014} a new activation function for RNNs was proposed.
The new activation function augments the usual logistic sigmoid activation
function with two gating units called reset, $\vr$, and update, $\vz$, gates.
Each gate depends on the previous hidden state $\vh^{(t-1)}$, and the current
input $\vx_t$ controls the flow of information. This is reminiscent of long
short-term memory (LSTM) units~\cite{Hochreiter1997}. For details about this
unit, we refer the reader to \cite{Cho2014} and Fig.~\ref{fig:rnn_unit}~(b). For
the remainder of this paper, we always use this new activation function.

\subsection{Gated Recursive Convolutional Neural Network}
\label{sec:grconv}

\begin{figure*}[ht]
    \centering
    \begin{minipage}{0.35\textwidth}
        \centering
        \includegraphics[width=0.8\columnwidth]{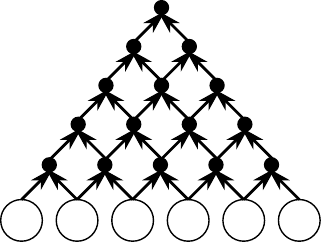}
    \end{minipage}
    \hfill
    \begin{minipage}{0.19\textwidth}
        \centering
        \includegraphics[width=0.8\columnwidth]{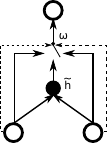}
    \end{minipage}
    \hfill
    \begin{minipage}{0.17\textwidth}
        \centering
        \includegraphics[width=0.8\columnwidth]{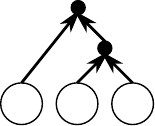}
    \end{minipage}
    \hfill
    \begin{minipage}{0.17\textwidth}
        \centering
        \includegraphics[width=0.8\columnwidth]{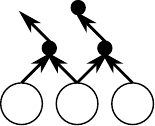}
    \end{minipage}
    \begin{minipage}{0.35\textwidth}
        \centering
        (a)
    \end{minipage}
    \hfill
    \begin{minipage}{0.19\textwidth}
        \centering
        (b)
    \end{minipage}
    \hfill
    \begin{minipage}{0.17\textwidth}
        \centering
        (c)
    \end{minipage}
    \hfill
    \begin{minipage}{0.17\textwidth}
        \centering
        (d)
    \end{minipage}
    \caption{The graphical illustration of (a) the recursive convolutional
        neural network and (b) the proposed gated unit for the
    recursive convolutional neural network. (c--d) The example structures that
may be learned with the proposed gated unit.}
    \label{fig:rconv_unit}
\end{figure*}

Besides RNNs, another natural approach to dealing with variable-length sequences
is to use a recursive convolutional neural network where the parameters at each
level are shared through the whole network (see Fig.~\ref{fig:rconv_unit}~(a)).
In this section, we introduce a binary convolutional neural network whose
weights are recursively applied to the input sequence until it outputs a single
fixed-length vector. In addition to a usual convolutional architecture, we
propose to use the previously mentioned gating mechanism, which allows the
recursive network to learn the structure of the source sentences on the fly.

Let $x=(\vx_1, \vx_2, \cdots, \vx_T)$ be an input sequence, where $\vx_t \in
\RR^d$.  The proposed gated recursive convolutional neural network (grConv)
consists of four weight matrices $\mW^l$, $\mW^r$, $\mG^l$ and $\mG^r$. At each
recursion level $t \in \left[ 1, T-1\right]$, the activation of the $j$-th
hidden unit $h^{(t)}_j$ is computed by
\begin{align}
    \label{eq:grconv_main}
    h^{(t)}_j = \omega_c \tilde{h}^{(t)}_j + \omega_l h^{(t-1)}_{j-1} + \omega_r
    h^{(t-1)}_j,
\end{align}
where $\omega_c$, $\omega_l$ and $\omega_r$ are the values of a gater that sum
to $1$. The hidden unit is initialized as
\begin{align*}
    h^{(0)}_j = \mU \vx_j,
\end{align*}
where $\mU$ projects the input into a hidden space.

The new activation $\tilde{h}^{(t)}_j$ is computed as usual:
\begin{align*}
    \tilde{h}^{(t)}_j = \phi\left( \mW^l h^{(t)}_{j-1} + \mW^r h^{(t)}_{j}
    \right),
\end{align*}
where $\phi$ is an element-wise nonlinearity.

The gating coefficients $\omega$'s are computed by
\begin{align*}
    \left[ 
        \begin{array}{c}
            \omega_c \\
            \omega_l \\
            \omega_r
        \end{array}
    \right] = \frac{1}{Z}
    \exp\left( \mG^l h^{(t)}_{j-1} + \mG^r h^{(t)}_{j}
    \right),
\end{align*}
where $\mG^l, \mG^r \in \RR^{3 \times d}$ and 
\[
    Z = \sum_{k=1}^3 \left[\exp\left( \mG^l h^{(t)}_{j-1} + \mG^r h^{(t)}_{j} \right)\right]_k.
\]

According to this activation, one can think of the activation of a single node
at recursion level $t$ as a choice between either a new activation computed from
both left and right children, the activation from the left child, or the
activation from the right child. This choice allows the overall structure of the
recursive convolution to change adaptively with respect to an input sample. See
Fig.~\ref{fig:rconv_unit}~(b) for an illustration.

In this respect, we may even consider the proposed grConv as doing a kind of
unsupervised parsing. If we consider the case where the gating unit makes a
hard decision, i.e., $\omega$ follows an 1-of-K coding, it is easy to see that
the network adapts to the input and forms a tree-like structure (See
Fig.~\ref{fig:rconv_unit}~(c--d)). However, we leave the further investigation
of the structure learned by this model for future research.

\section{Purely Neural Machine Translation}

\subsection{Encoder--Decoder Approach}

The task of translation can be understood from the perspective of machine learning
as learning the conditional distribution $p(f \mid e)$ of a target sentence
(translation) $f$ given a source sentence $e$. Once the conditional distribution
is learned by a model, one can use the model to directly sample a target
sentence given a source sentence, either by actual sampling or by using a
(approximate) search algorithm to find the maximum of the distribution.

A number of recent papers have proposed to use neural networks to directly learn
the conditional distribution from a bilingual, parallel
corpus~\cite{Kalchbrenner2012,Cho2014,Sutskever2014}. For instance, the authors
of \cite{Kalchbrenner2012} proposed an approach involving a convolutional
$n$-gram model to extract a fixed-length vector of a source sentence which is
decoded with an inverse convolutional $n$-gram model augmented with an RNN. In
\cite{Sutskever2014}, an RNN with LSTM units was used to encode a source
sentence and starting from the last hidden state, to decode a target sentence.
Similarly, the authors of \cite{Cho2014} proposed to use an RNN to encode and
decode a pair of source and target phrases.

\begin{figure}
    \centering
    \includegraphics[width=0.9\columnwidth]{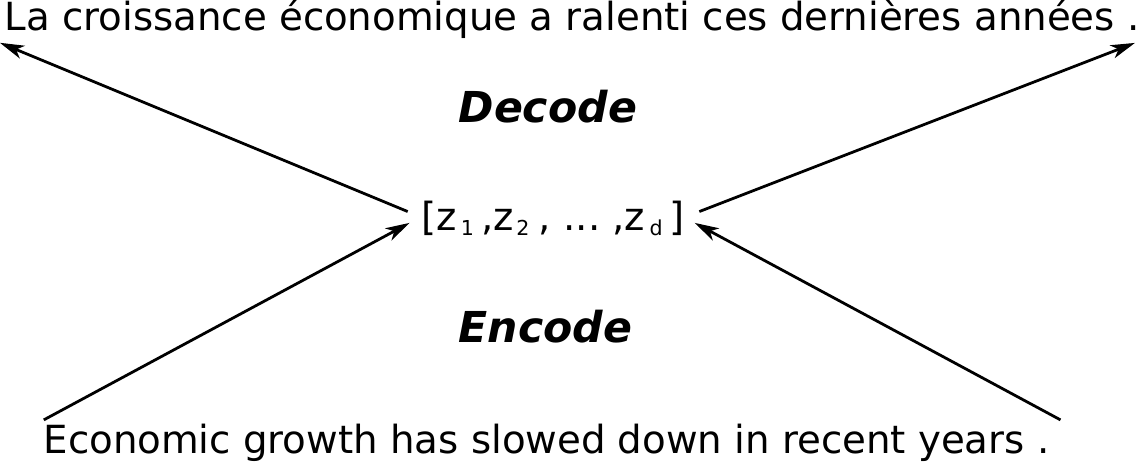}
    \caption{The encoder--decoder architecture} 
    \label{fig:encode_decode}
\end{figure}

At the core of all these recent works lies an encoder--decoder architecture (see
Fig.~\ref{fig:encode_decode}). The encoder processes a variable-length input
(source sentence) and builds a fixed-length vector representation (denoted as
$\vz$ in Fig.~\ref{fig:encode_decode}). Conditioned on the encoded
representation, the decoder generates a variable-length sequence (target
sentence).

Before \cite{Sutskever2014} this encoder--decoder approach was used mainly as
a part of the existing statistical machine translation (SMT) system. This
approach was used to re-rank the $n$-best list generated by the SMT system in
\cite{Kalchbrenner2012}, and the authors of \cite{Cho2014} used this approach
to provide an additional score for the existing phrase table.

In this paper, we concentrate on analyzing the direct translation performance,
as in \cite{Sutskever2014}, with two model configurations. In both models, we
use an RNN with the gated hidden unit~\cite{Cho2014}, as this is one of the only
options that does not require a non-trivial way to determine the target length.
The first model will use the same RNN with the gated hidden unit as an encoder,
as in \cite{Cho2014}, and the second one will use the proposed gated recursive
convolutional neural network (grConv). We aim to understand the inductive bias
of the encoder--decoder approach on the translation performance measured by
BLEU.

\section{Experiment Settings}

\subsection{Dataset}

We evaluate the encoder--decoder models on the task of English-to-French
translation. We use the bilingual, parallel corpus which is a set of 348M
selected by the method in \cite{Axelrod2011} from a combination of Europarl (61M
words), news commentary (5.5M), UN (421M) and two crawled corpora of 90M and
780M words respectively.\footnote{All the data can be downloaded from
\url{http://www-lium.univ-lemans.fr/~schwenk/cslm_joint_paper/}.} We did not use
separate monolingual data. The performance of the neural machien translation
models was measured on the news-test2012, news-test2013 and news-test2014 sets
(~3000 lines each). When comparing to the SMT system, we use news-test2012 and
news-test2013 as our development set for tuning the SMT system, and
news-test2014 as our test set.

Among all the sentence pairs in the prepared parallel corpus, for reasons of
computational efficiency  we only use the pairs where both English and French
sentences are at most 30 words long to train neural networks. Furthermore, we
use only the 30,000 most frequent words for both English and French. All the
other rare words are considered unknown and are mapped to a special token
($\left[ \text{UNK} \right]$).

\subsection{Models}


We train two models: The RNN Encoder--Decoder~(RNNenc)\cite{Cho2014} and the
newly proposed gated recursive convolutional neural network (grConv). Note that
both models use an RNN with gated hidden units as a decoder (see
Sec.~\ref{sec:rnn_gated}).

We use minibatch stochastic gradient descent with AdaDelta~\cite{Zeiler-2012} to
train our two models. We initialize the square weight matrix (transition matrix)
as an orthogonal matrix with its spectral radius set to $1$ in the case of the
RNNenc and $0.4$ in the case of the grConv. $\tanh$ and a rectifier
($\max(0,x)$) are used as the element-wise nonlinear functions for the RNNenc
and grConv respectively.

The grConv has 2000 hidden neurons, whereas the RNNenc has 1000 hidden
neurons. The word embeddings are 620-dimensional in both cases.\footnote{
  In all cases, we train the whole network including the word embedding matrix. 
}
Both models were trained for approximately 110 hours, which is equivalent to
296,144 updates and 846,322 updates for the grConv and RNNenc,
respectively.\footnote{
  The code will be available online, should the paper be accepted.
}

\begin{table*}[ht]
    \centering
    \begin{minipage}{0.48\textwidth}
    \centering
    \begin{tabular}{c | c | c c}
        & Model & Development & Test \\
        \hline
        \hline
        \multirow{5}{*}{\rotatebox[origin=c]{90}{All}} 
        & RNNenc & 13.15 & 13.92 \\
        & grConv & 9.97 & 9.97 \\
        & Moses  & 30.64 & 33.30 \\
        & Moses+RNNenc$^\star$ & 31.48 & 34.64 \\
        & Moses+LSTM$^\circ$ & 32 & 35.65 \\
        \hline
        \multirow{3}{*}{\rotatebox[origin=c]{90}{No UNK}}
        & RNNenc & 21.01 & 23.45 \\
        & grConv & 17.19 & 18.22 \\
        & Moses  & 32.77 & 35.63 \\
    \end{tabular}
\end{minipage}
\hfill
    \begin{minipage}{0.48\textwidth}
    \centering
    \begin{tabular}{c | c | c c}
        & Model & Development & Test \\
        \hline
        \hline
        \multirow{3}{*}{\rotatebox[origin=c]{90}{All}} 
        & RNNenc & 19.12 & 20.99 \\
        & grConv & 16.60 & 17.50 \\
        & Moses  & 28.92 & 32.00 \\
        \hline
        \multirow{3}{*}{\rotatebox[origin=c]{90}{No UNK}}
        & RNNenc & 24.73 & 27.03 \\
        & grConv & 21.74 & 22.94 \\
        & Moses  & 32.20 & 35.40 \\
    \end{tabular}
\end{minipage}

    \begin{minipage}{0.48\textwidth}
    \centering
    (a) All Lengths
\end{minipage}
\hfill
    \begin{minipage}{0.48\textwidth}
    \centering
    (b) 10--20 Words
\end{minipage}
    \caption{BLEU scores computed on the development and test sets. The top
        three rows show the scores on all the sentences, and the bottom three
        rows on the sentences having no unknown words. ($\star$) The result
        reported in \cite{Cho2014} where the RNNenc was used to score phrase
        pairs in the phrase table. ($\circ$) The result reported in
        \cite{Sutskever2014} where an encoder--decoder with LSTM units was used
    to re-rank the $n$-best list generated by Moses.}
    \label{tab:bleu}
\end{table*}

\subsubsection{Translation using Beam-Search}

We use a basic form of beam-search to find a translation that maximizes the
conditional probability given by a specific model (in this case, either the
RNNenc or the grConv). At each time step of the decoder, we keep the $s$
translation candidates with the highest log-probability, where $s=10$ is the
beam-width.  During the beam-search, we exclude any hypothesis that includes an
unknown word.  For each end-of-sequence symbol that is selected among the
highest scoring candidates the beam-width is reduced by one, until the
beam-width reaches zero.

The beam-search to (approximately) find a sequence of maximum log-probability
under RNN was proposed and used successfully in \cite{Graves2012} and
\cite{Boulanger2013}.  Recently, the authors of \cite{Sutskever2014} found this
approach to be effective in purely neural machine translation based on LSTM
units.

When we use the beam-search to find the $k$ best translations, we do not use a
usual log-probability but one normalized with respect to the length of the
translation.  This prevents the RNN decoder from favoring shorter translations,
behavior which was observed earlier in, e.g.,~\cite{Graves2013}.

\begin{figure*}[ht]
  \centering
  \begin{minipage}{0.31\textwidth}
      \centering
      \includegraphics[width=1.\columnwidth]{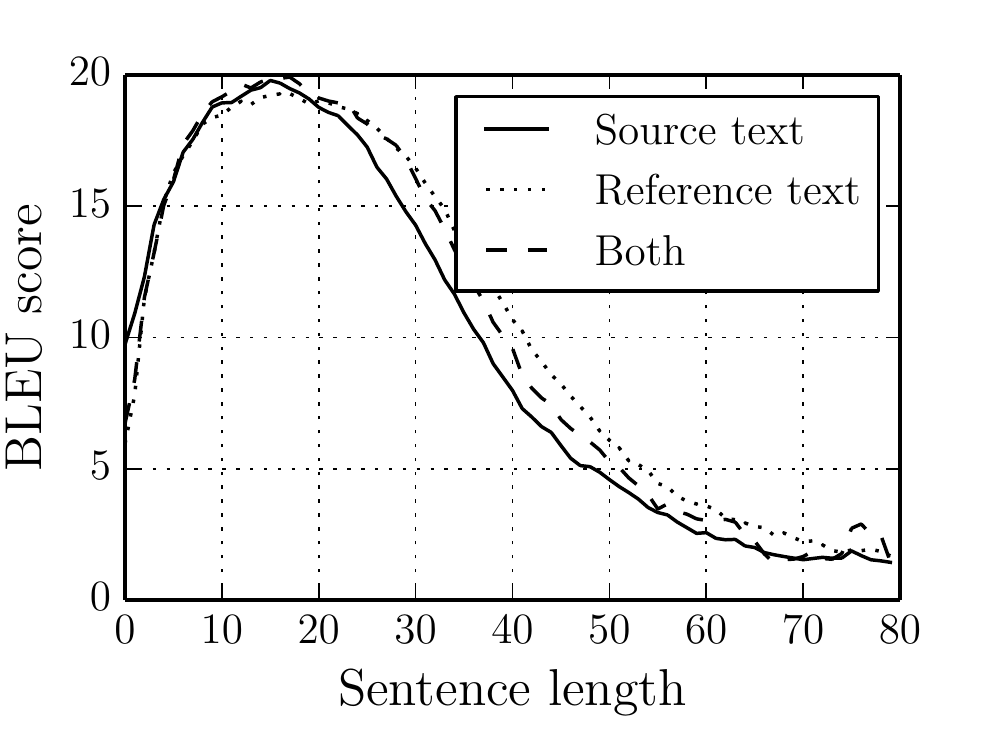}
      \\
      (a) RNNenc
  \end{minipage}
  \hfill
  \begin{minipage}{0.31\textwidth}
      \centering
      \includegraphics[width=1.\columnwidth]{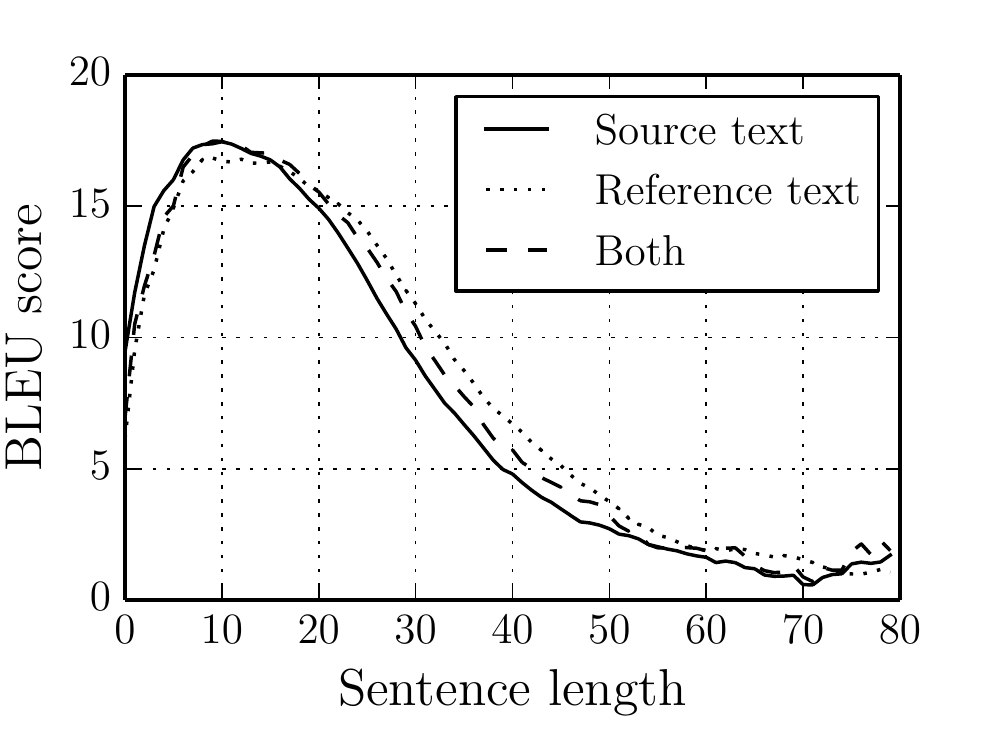}
      \\
      (b) grConv
  \end{minipage}
  \hfill
  \begin{minipage}{0.31\textwidth}
      \centering
      \includegraphics[width=1.\columnwidth]{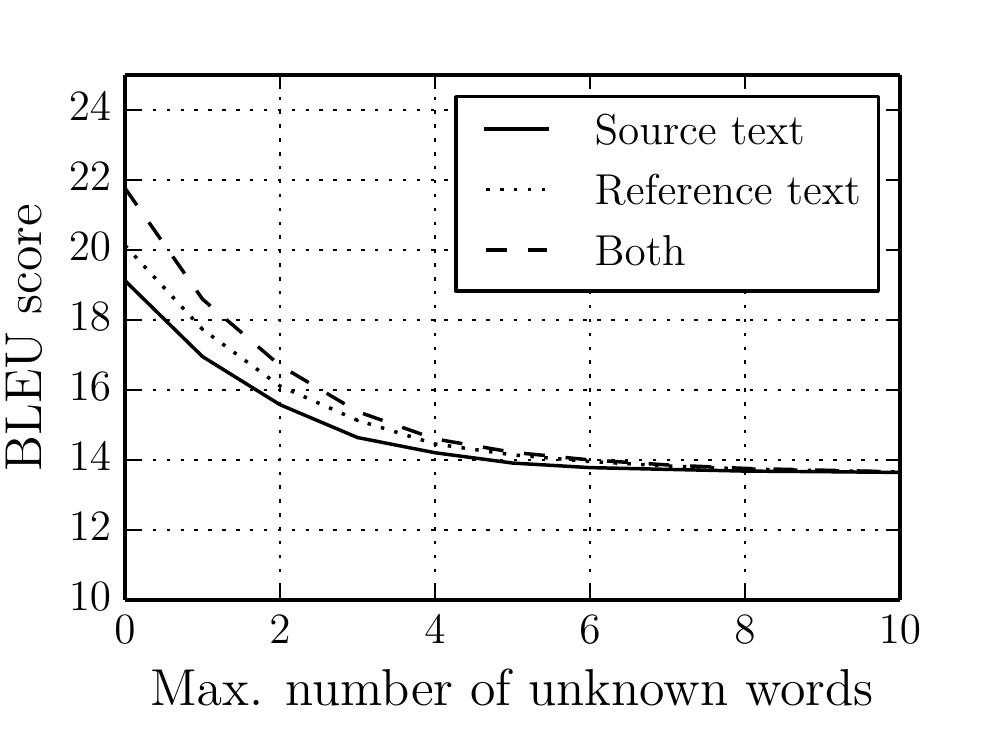}
      \\
      (c) RNNenc
  \end{minipage}
  \caption{The BLEU scores achieved by (a) the RNNenc and (b) the grConv for
  sentences of a given length. The plot is smoothed by taking a window of size
  10. (c) The BLEU scores achieved by the RNN model for sentences with less
  than a given number of unknown words.}
  \label{fig:bleu_length}
\end{figure*}

\begin{table*}[htp]
    \begin{minipage}{0.99\textwidth}
        \small
        \centering
        \begin{tabular}{c | p{13cm}}
Source & She explained her new position of foreign affairs and security policy
representative as a reply to a question: "Who is the European Union? Which phone
number should I call?"; i.e. as an important step to unification and better
clarity of Union's policy towards countries such as China or India. \\
            \hline
Reference & Elle a expliqué le nouveau poste de la Haute représentante pour les
affaires étrangères et la politique de défense dans le cadre d'une réponse à la
question: "Qui est qui à l'Union européenne?" "A quel numéro de téléphone
dois-je appeler?", donc comme un pas important vers l'unicité et une plus grande
lisibilité de la politique de l'Union face aux états, comme est la Chine ou bien
l'Inde. \\
            \hline
RNNEnc & Elle a décrit sa position en matière de politique étrangère et de
sécurité ainsi que la politique de l'Union européenne en matière de gouvernance
et de démocratie . \\
            \hline
grConv & Elle a expliqué sa nouvelle politique étrangère et de sécurité en
réponse à un certain nombre de questions : "Qu'est-ce que l'Union européenne ? "
. \\
            \hline
Moses & Elle a expliqué son nouveau poste des affaires étrangères et la
politique de sécurité représentant en réponse à une question: "Qui est l'Union
européenne? Quel numéro de téléphone dois-je appeler?"; c'est comme une étape
importante de l'unification et une meilleure lisibilité de la politique de
l'Union à des pays comme la Chine ou l'Inde . \\
            \hline
\multicolumn{2}{c}{} \\
\hline
Source & The investigation should be complete by the end of the year when the
findings will be presented to Deutsche Bank's board of managing directors - with
recommendations for action. \\
            \hline
Reference & L'examen doit être terminé d'ici la fin de l'année, ensuite les
résultats du conseil d'administration de la Deutsche Bank doivent être présentés
- avec recommandation, d' habitude. \\
            \hline
RNNEnc & L'étude devrait être terminée à la fin de l' année, lorsque les
conclusions seront présentées au conseil d'administration de la Deutsche Bank,
conseil d'association avec des mesures. \\
            \hline
grConv & L'enquête devrait être terminée à la fin de l'année où les conclusions
seront présentées par le conseil d'administration de la BCE à la direction des
recommandations. \\
            \hline
Moses & L'enquête devrait être terminé d'ici la fin de l'année lorsque les
résultats seront présentés à la Deutsche Bank conseil des directeurs généraux -
avec des recommandations . \\
\hline
\multicolumn{2}{c}{} \\
\hline
Source &  And there are thorny mechanical questions that must be resolved during
that time, like how to balance the state's mandate of "adequate access" to
licensed marijuana with its prohibitions on cannabis businesses within 1,000
feet of a school, park, playground or child care center. \\
          \hline
Reference &  Pendant ce temps, des questions pratiques restent en suspens:
comment équilibrer le mandat de l'état qui garantit un accès approprié à la
marijuana agréée et interdit l'installation de commerces de vente de cannabis
dans un rayon de 30 km autour d'une école, d'un parc, d'un terrain de jeu ou
d'une crèche. \\
            \hline
RNNEnc &  Il y a des questions préventives qui se posent quant à l'équilibre des
droits de l'enfant dans les limites d'une entreprise de collecte de sang. \\
            \hline
grConv &  De façon générale, il y a des raisons de sécurité pour que les
entreprises aient accès à des milliers de centres de pêche, d'eau ou de
recherche. \\
            \hline
Moses &  Et il y a des problèmes mécaniques complexes qui doivent être résolues
au cours de cette période, comme la manière d'équilibrer le mandat de "l'accès
adéquat" permis de marijuana avec l'interdiction du cannabis aux entreprises de
1000 pieds d'une école de jeu ou de parc, le service de garde. \\
        \end{tabular}
        \\
        {\normalsize (a) Long Sentences}
    \end{minipage}

    \vspace{3mm}
    \begin{minipage}{0.99\textwidth}
        \small
        \centering
        \begin{tabular}{c | p{13cm}}
Source & There is still no agreement as to which election rules to follow. \\
            \hline
Reference & Toutefois il n'existe toujours pas d'accord selon quel règlement de
vote il faut procéder.  \\
            \hline
RNNEnc & Il n'y a pas encore d'accord sur les règles électorales. \\
            \hline
grConv & Il n'y a pas encore d'accord sur la question des élections à suivre. \\
            \hline
Moses & Il y a toujours pas d'accord sur l'élection des règles à suivre. \\
\hline
\multicolumn{2}{c}{} \\
\hline
Source & Many of these ideas may have been creative, but they didn't necessarily
work. \\
            \hline
Reference & Beaucoup de ces idées étaient créatives mais elles n'ont pas
forcément fonctionné. \\
            \hline
RNNEnc & Bon nombre de ces idées ont peut-être été créatrices, mais elles ne
s'appliquaient pas nécessairement. \\
            \hline
grConv & Beaucoup de ces idées peuvent être créatives, mais elles n'ont pas
fonctionné. \\
            \hline
Moses & Beaucoup de ces idées ont pu être créatif, mais ils n'ont pas
nécessairement. \\
\hline
\multicolumn{2}{c}{} \\
\hline
Source & There is a lot of consensus between the Left and the Right on this
subject. \\
            \hline
Reference & C'est qu'il y a sur ce sujet un assez large consensus entre gauche
et droite. \\
            \hline
RNNEnc & Il existe beaucoup de consensus entre la gauche et le droit à la
question. \\
            \hline
grConv & Il y a un consensus entre la gauche et le droit sur cette question. \\
            \hline
Moses & Il y a beaucoup de consensus entre la gauche et la droite sur ce sujet.
\\
\hline
\multicolumn{2}{c}{} \\
\hline
Source & According to them, one can find any weapon at a low price right now. \\
            \hline
Reference & Selon eux, on peut trouver aujourd'hui à Moscou n'importe quelle
arme pour un prix raisonnable.\\
            \hline
RNNEnc & Selon eux, on peut se trouver de l'arme à un prix trop bas.\\
            \hline
grConv & En tout cas, ils peuvent trouver une arme à un prix très bas à la
fois.\\
            \hline
Moses & Selon eux, on trouve une arme à bas prix pour l'instant.
        \end{tabular}
        \\
        {\normalsize (b) Short Sentences}
    \end{minipage}
    \caption{The sample translations along with the source sentences and the reference translations.} 
    \label{tbl:translations}
\end{table*}

\section{Results and Analysis}

\subsection{Quantitative Analysis}

In this paper, we are interested in the properties of the neural machine
translation models. Specifically, the translation quality with respect to the
length of source and/or target sentences and with respect to the number of words
unknown to the model in each source/target sentence.

First, we look at how the BLEU score, reflecting the translation performance,
changes with respect to the length of the sentences (see
Fig.~\ref{fig:bleu_length} (a)--(b)). Clearly, both models perform relatively
well on short sentences, but suffer significantly as the length of the
sentences increases.

We observe a similar trend with the number of unknown words, in
Fig.~\ref{fig:bleu_length} (c). As expected, the performance degrades rapidly as
the number of unknown words increases. This suggests that it will be an
important challenge to increase the size of vocabularies used by the neural
machine translation system in the future. Although we only present the result
with the RNNenc, we observed similar behavior for the grConv as well.

In Table~\ref{tab:bleu}~(a), we present the translation performances obtained
using the two models along with the baseline phrase-based SMT system.\footnote{
    We used Moses as a baseline, trained with additional monolingual data for a
    4-gram language model.
} Clearly the phrase-based SMT system still shows the superior performance over
the proposed purely neural machine translation system, but we can see that under
certain conditions (no unknown words in both source and reference sentences),
the difference diminishes quite significantly. Furthermore, if we consider only
short sentences (10--20 words per sentence), the difference further decreases
(see Table~\ref{tab:bleu}~(b). 

Furthermore, it is possible to use the neural machine translation models
together with the existing phrase-based system, which was found recently in
\cite{Cho2014,Sutskever2014} to improve the overall translation performance
(see Table~\ref{tab:bleu}~(a)).

This analysis suggests that that the current neural translation approach has
its weakness in handling long sentences. The most obvious explanatory
hypothesis is that the fixed-length vector representation does not have enough
capacity to encode a long sentence with complicated structure and meaning. In
order to encode a variable-length sequence, a neural network may ``sacrifice''
some of the important topics in the input sentence in order to remember others.

\begin{figure}[ht]
  \centering
  \includegraphics[width=0.9\columnwidth]{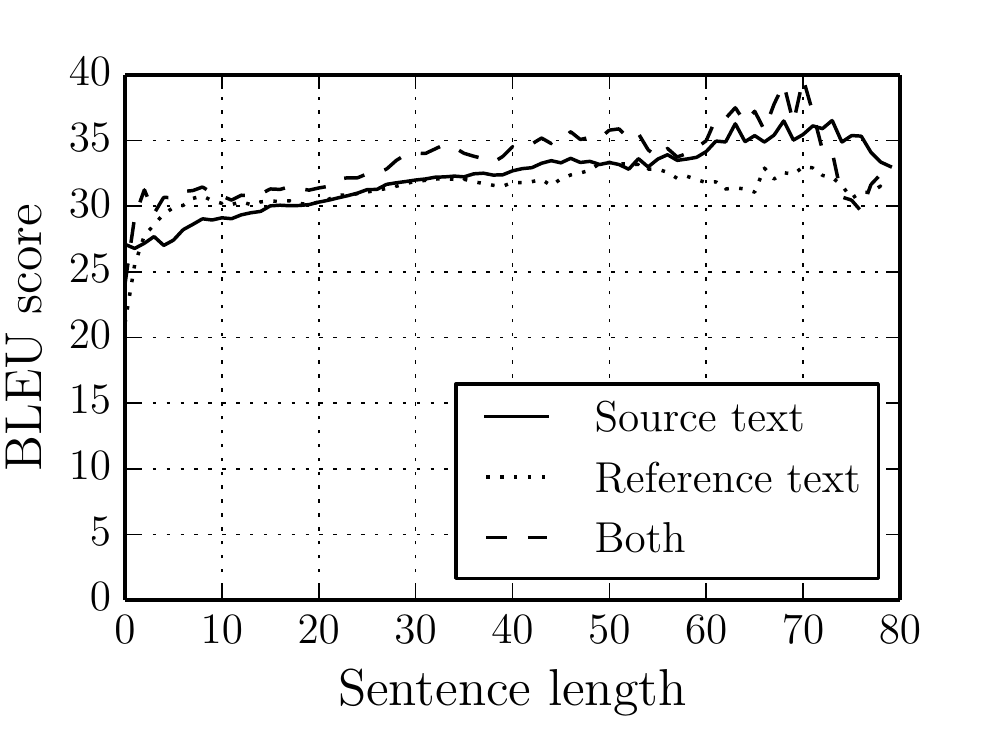} \caption{The BLEU
  scores achieved by an SMT system for sentences of a given length. The plot is
  smoothed by taking a window of size 10.
  We use the solid, dotted and dashed
  lines to show the effect of different lengths of source, reference or both
  of them, respectively.}
  \label{fig:moses_bleu_length}
\end{figure}

This is in stark contrast to the conventional phrase-based machine translation
system~\cite{Koehn2003}. As we can see from Fig.~\ref{fig:moses_bleu_length},
the conventional system trained on the same dataset (with additional monolingual
data for the language model) tends to get a higher BLEU score on longer
sentences.

In fact, if we limit the lengths of both the source sentence and the reference
translation to be between 10 and 20 words and use only the sentences with no
unknown words, the BLEU scores on the test set are 27.81 and 33.08 for the
RNNenc and Moses, respectively.

Note that we observed a similar trend even when we used sentences of up to 50
words to train these models.

\subsection{Qualitative Analysis}

Although BLEU score is used as a de-facto standard metric for evaluating the
performance of a machine translation system, it is not the perfect metric~(see,
e.g., \cite{Song13,Liu2011}). Hence, here we present some of the actual
translations generated from the two models, RNNenc and grConv.

In Table.~\ref{tbl:translations} (a)--(b), we show the translations of some
randomly selected sentences from the development and test sets. We chose the
ones that have no unknown words. (a) lists long sentences (longer than 30
words), and (b) short sentences (shorter than 10 words). We can see that,
despite the difference in the BLEU scores, all three models (RNNenc, grConv and
Moses) do a decent job at translating, especially, short sentences. When the
source sentences are long, however, we notice the performance degradation of the
neural machine translation models.

\begin{figure*}[ht]
    \begin{minipage}{0.5\textwidth}
        \centering
        \includegraphics[width=\textwidth,clip=true,trim=90 50 90 50]{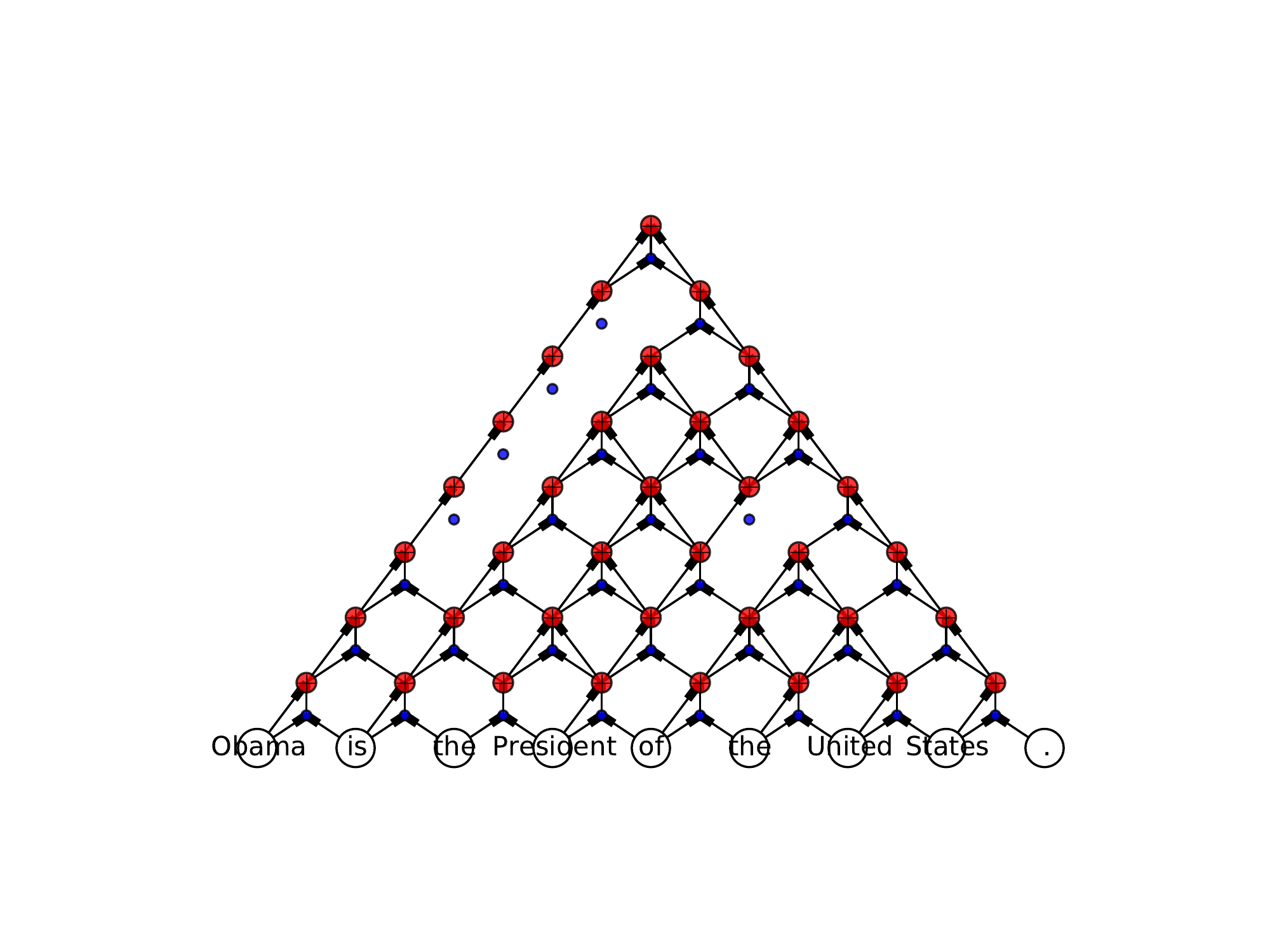}
    \end{minipage}
    \hfill
    \begin{minipage}{0.48\textwidth}
        \centering
        \begin{tabular}{l}
            Translations \\
            \hline
Obama est le Président des États-Unis .        (2.06)\\
Obama est le président des États-Unis .        (2.09)\\
Obama est le président des Etats-Unis .        (2.61)\\
Obama est le Président des Etats-Unis .        (3.33)\\
Barack Obama est le président des États-Unis . (4.41)\\
Barack Obama est le Président des États-Unis . (4.48)\\
Barack Obama est le président des Etats-Unis . (4.54)\\
L'Obama est le Président des États-Unis .      (4.59)\\
L'Obama est le président des États-Unis .      (4.67)\\
Obama est président du Congrès des États-Unis .(5.09) \\
        \end{tabular}
    \end{minipage}
    \begin{minipage}{0.4\textwidth}
        \centering
        (a)
    \end{minipage}
    \hfill
    \begin{minipage}{0.58\textwidth}
        \centering
        (b)
    \end{minipage}
    \caption{(a) The visualization of the grConv structure when the input is
        {\it ``Obama is the President of the United States.''}. Only edges with
        gating coefficient $\omega$ higher than $0.1$ are shown. (b) The top-$10$ translations
        generated by the grConv. The numbers in parentheses are the negative
    log-probability of the translations.}
    \label{fig:obama}
\end{figure*}

Additionally, we present here what type of structure the proposed gated
recursive convolutional network learns to represent. With a sample sentence
{\it ``Obama is the President of the United States''}, we present the parsing
structure learned by the grConv encoder and the generated translations, in
Fig.~\ref{fig:obama}. The figure suggests that the grConv extracts the vector
representation of the sentence by first merging {\it ``of the United States''}
together with {\it ``is the President of''} and finally combining this with {\it
``Obama is''} and {\it ``.''}, which is well correlated with our intuition.

Despite the lower performance the grConv showed compared
to the RNN Encoder--Decoder,\footnote{
    However, it should be noted that the number of gradient updates used to
    train the grConv was a third of that used to train the RNNenc. Longer
    training may change the result, but for a fair comparison we chose to
    compare models which were trained for an equal amount of time. Neither model
    was trained to convergence.
}
we find this property of the grConv learning a
grammar structure automatically interesting and believe further investigation is
needed.

\section{Conclusion and Discussion}

In this paper, we have investigated the property of a recently introduced family
of machine translation system based purely on neural networks. We focused on
evaluating an encoder--decoder approach, proposed recently in
\cite{Kalchbrenner2012,Cho2014,Sutskever2014}, on the task of
sentence-to-sentence translation. Among many possible encoder--decoder models we
specifically chose two models that differ in the choice of the encoder; (1) RNN
with gated hidden units and (2) the newly proposed gated recursive convolutional
neural network.

After training those two models on pairs of English and French sentences, we
analyzed their performance using BLEU scores with respect to the lengths of
sentences and the existence of unknown/rare words in sentences. Our analysis
revealed that the performance of the neural machine translation suffers
significantly from the length of sentences. However, qualitatively, we found
that the both models are able to generate correct translations very well.

These analyses suggest a number of future research directions in machine
translation purely based on neural networks.

Firstly, it is important to find a way to scale up training a neural network
both in terms of computation and memory so that much larger vocabularies for
both source and target languages can be used. Especially, when it comes to
languages with rich morphology, we may be required to come up with a radically
different approach in dealing with words.

Secondly, more research is needed to prevent the neural machine translation
system from underperforming with long sentences.  Lastly, we need to explore
different neural architectures, especially for the decoder. Despite the radical
difference in the architecture between RNN and grConv which were used as an
encoder, both models suffer from {\it the curse of sentence length}. This
suggests that it may be due to the lack of representational power in the
decoder. Further investigation and research are required.

In addition to the property of a general neural machine translation system, we
observed one interesting property of the proposed gated recursive convolutional
neural network (grConv). The grConv was found to mimic the grammatical structure
of an input sentence without any supervision on syntactic structure of language.
We believe this property makes it appropriate for natural language processing
applications other than machine translation.

\section*{Acknowledgments}

The authors would like to acknowledge the support of the following agencies for
research funding and computing support: NSERC, Calcul Qu\'{e}bec, Compute Canada,
the Canada Research Chairs and CIFAR.

\bibliographystyle{acl}
\bibliography{strings,strings-shorter,ml,aigaion,myref}

\end{document}